\documentclass[USenglish,twocolumn]{article}
\usepackage[utf8]{inputenc}
\usepackage[big,online]{dgruyter}
\usepackage[sort&compress,square,numbers]{natbib}
\usepackage{times}

\begin{document}

  \DOI{10.1515/}
  \openaccess
  \pagenumbering{gobble}

\title{HOLa: HoloLens Object Labeling}
\runningtitle{HOLa: HoloLens Object Labeling}

\author*[1]{Michael Schwimmbeck}
\author[2]{Serouj Khajarian}
\author[3]{Konstantin Holzapfel}
\author[3]{Johannes Schmidt}
\author[2]{Stefanie Remmele} 
\runningauthor{M.~Schwimmbeck et al.}

\affil[1]{\protect\raggedright 
  University of Applied Sciences Landshut, Am Lurzenhof 1, 84036 Landshut, Germany, e-mail: s-mschw8@haw-landshut.de}
\affil[2]{\protect\raggedright
  University of Applied Sciences Landshut, Landshut, Germany}
\affil[3]{\protect\raggedright
  LAKUMED Hospital Landshut-Achdorf, Achdorfer Weg 3, 84036 Landshut, Germany}

\abstract{In the context of medical Augmented Reality (AR) applications, object tracking is a key challenge and requires a significant amount of annotation masks. As segmentation foundation models like the Segment Anything Model (SAM) begin to emerge, zero-shot segmentation requires only minimal human participation obtaining high-quality object masks.
We introduce a HoloLens-Object-Labeling (HOLa) Unity and Python application based on the SAM-Track algorithm that offers fully automatic single object annotation for HoloLens 2 while requiring minimal human participation. HOLa does not have to be adjusted to a specific image appearance and could thus alleviate AR research in any application field.
We evaluate HOLa for different degrees of image complexity in open liver surgery and in medical phantom experiments. Using HOLa for image annotation can increase the labeling speed by more than 500 times while providing Dice scores between 0.875 and 0.982, which are comparable to human annotators. Our code is publicly available at: \url{https://github.com/mschwimmbeck/HOLa}}


\maketitle

\section{Introduction} 

In AR-guided surgery, organ segmentation or tracking and virtual model registration are important tasks, typically solved by deep learning solutions \citep{1}. Training task-specific models has been the standard for a long time, resulting in restricted model applicability. However, data annotation or pixel-wise image labeling is known to be expensive. Recently, Kirillov et al. \citep{2} introduced a task-independent foundation model approach by introducing the Segment Anything Model (SAM). The use of SAM creates manifold applications in computer vision and beyond. Furthermore, SAM offers promptable segmentation allowing for user constraints for more demanding tasks \citep{2,3}.

Although SAM makes data annotation less expensive, annotating whole sequences, for instance all frames in a video, remains time consuming since it does not exploit the information redundancy of consecutive frames. Therefore, some first approaches propose to combine SAM with tracking algorithms to track initially segmented objects throughout the whole sequence \citep{3}. SAM-Track by Cheng et al. \citep{3} for instance offers interactive and automatic methods for offline segmentation and tracking of objects in pre-recorded videos utilizing DeAOT tracker \citep{4}.

The majority of all AR work using optical see-through head mounted displays (OST-HMDs) is conducted with the Microsoft HoloLens \citep{1}. Further, the HoloLens is one of the most commonly used OST-HMDs in medical AR studies \citep{1}. To establish pixel-wise tracking models for the field, it would be more convenient to minimize the manual labeling effort for training data by handing it over to a sequence labeling tool like SAM-Track. This way, the user is only required to initialize SAM-Track correctly once and delegate the annotation task to the algorithm.

The main contributions of this work are first, the investigation of the SAM-Track method for the labeling of HoloLens RGB data and second, the integration of the application in a HoloLens recording app to simplify data management during development of deep learning based tracking algorithms. We provide our recording app via the HoloLens-Object-Labeling (HOLa) application.

\section{Methods} 

\begin{figure}[!htb]
	\includegraphics[width=\columnwidth]{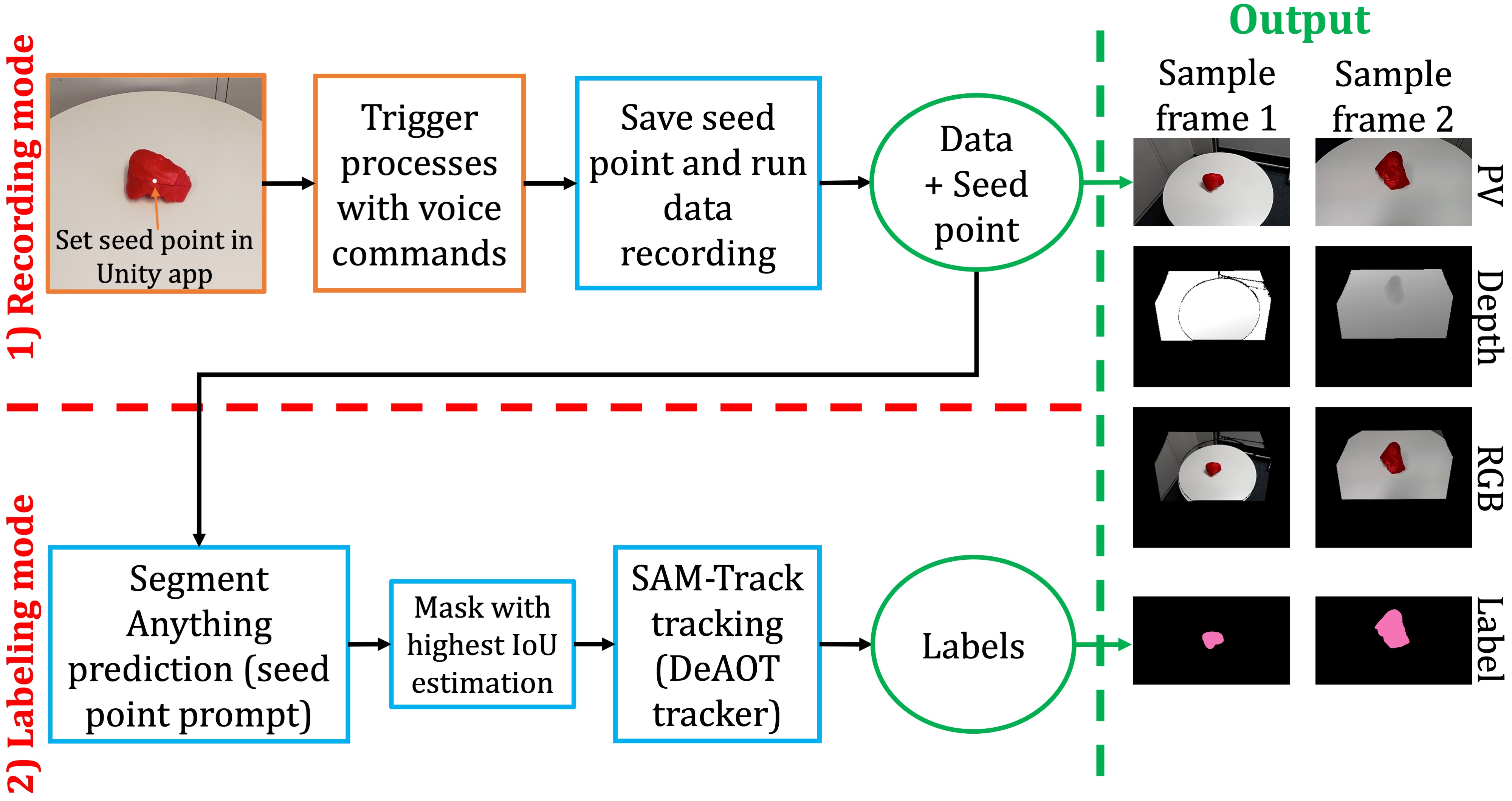}
	\caption{The HOLa app consists of two modes. In the recording mode, first, the user points a sphere cursor onto the object of interest by head motion. He selects the cursor position as seed point for segmentation by voice command, which also starts sensor recording. The labeling mode utilizes the SAM \citep{2} seed point prompt to initialize the SAM-Track \citep{3} DeAOT tracker \citep{4} that tracks the segmented object throughout all subsequent frames to obtain pixel-wise labels.}
	\label{img:Figure1}
\end{figure}

The HOLa app is developed in Unity and Python and consists of a recording mode and a labeling mode. A general overview of the HOLa components and the two HOLa modes are shown in Figure \ref{img:Figure1}.\\
\\
\textbf{Recording Mode:} The HOLa Unity app uses the HL2SS plugin for Unity \citep{5} to stream all data via WLAN to a PC for recording. Hence, HOLa acquires the streams of the RGB camera in 640x360 pixels resolution (PV) and transformed to the depth sensor grid (RGB), the depth stream (D), the generated point cloud (PC) and camera poses. SAM-Track \citep{3} requires a seed point within the object of interest. We use a sphere cursor visualized in AR indicating the center of the recorded frame. The user positions that cursor in the center of the target object by head motion and saves this initial seed point position by voice command ("Start"). In response, the cursor changes its color and is deactivated, the seed point is saved and recording starts. Another voice command ("Stop") terminates recording, while the sphere color switches again to provide visual feedback to the user. To implement communication between Python and Unity scripts, we use the HL2SS \citep{5} IPC port.\\
\\
\textbf{Labeling Mode:} The labeling mode performs pixel-wise labeling of all RGB frames (PV) and delivers annotations in the PV resolution of 640x360 pixels. By default, SAM performs internal rescaling and padding to its input resolution of 1024x1024 pixels \citep{2}. HOLa is built on SAM-Track with a ViT-H SAM model. We transform all recorded PV frames to a video prior to frame-wise labeling. Furthermore, we modify SAM-Track tracking settings to track only one single object and to ignore other objects appearing at subsequent frames. Additionally, SAM-Track ”Segment Everything” mode is replaced with a seed point-prompted SAM application \citep{2}. Thus, the frame center point is used as seed point prompt in the first frame to initialize SAM-Track with one single object mask. When handing over a seed point prompt to SAM, three mask proposals including an internal rating are generated to handle ambiguity \citep{2}. The mask with the highest IoU score initializes the tracking part of SAM-Track that segments the object in every downstream frame. Eventually, an annotated mask of the cursor-marked object of interest is generated for each recorded PV frame.

\section{Experiments}

We conduct three phantom experiments and two clinical experiments evaluating HOLa labeling quality for different degrees of image complexity with focus on open liver surgery. Experiment 1-3 evaluate the HOLa performance on different phantoms. Even though phantom experiments simplify the labeling task compared to clinical experiments, phantom experiments are highly relevant in medical AR research \citep{1}. Hence, Experiment 1 covers a uni-color 3D-printed replication of a human liver, Experiment 2 a simple 3D-printed replication of a human hip bone in a low-contrast scenario, and Experiment 3 a liver phantom embedded into a torso phantom. Experiment 4 and 5 evaluate the HOLa performance on different clinical cases by the annotation of a real human liver in-situ. These experiments differ in image complexity. While the liver in experiment 4 is clearly separated from its surroundings, experiment 5 shows a liver that is strongly embedded in its surroundings with shallow transitions between the structures.

We recorded frames from all phantoms and clinical scenarios directly prior to resection. Subsequently, 90 frames were selected across the entire sequence that best represent the variations during recording. No frames were excluded in advance. Distances and viewpoints were chosen to produce scenes, which are comparable to possible surgeon‘s scenes during operation. The reference rater (HA 1) segmented the selected 90 frames of each experiment to be compared against the labeling performance of the algorithm. HA 1 is the same person who recorded the data and is thus most familiar with the object geometries. To rate annotation quality, we computed an averaged Dice score over the 90 selected frames using the reference annotation (HA 1) as ground-truth. To ensure label quality, the ground-truth labels were revised by two medical experts (17 and 35 years of experience).

Additionally, to compare the discrepancy of man and machine labels against human discrepancies, we chose 10 representative frames from each experiment that were given to four additional annotators (HA 2-5) with technical background instructed by medical doctors on example images. Thus, we analyze inter-rater concordance and further validate the human reference as ground truth (HA 1). For the selected 10 frames, we compare Dice scores a) between the reference annotations of HA 1 and the HOLa annotations as well as b) between HA 1 and each of the four additional annotators (on average).

Furthermore, to quantify the annotation cost reduction, we compare the mean labeling speed per frame between HOLa and humans in all five experiments. An NVIDIA GeForce RTX 3090 GPU was used for HOLa post processing. For reference annotation, we used MATLAB Image Labeler (The MathWorks Inc.).

\section{Results}

\begin{figure}
	\includegraphics[width=\columnwidth]{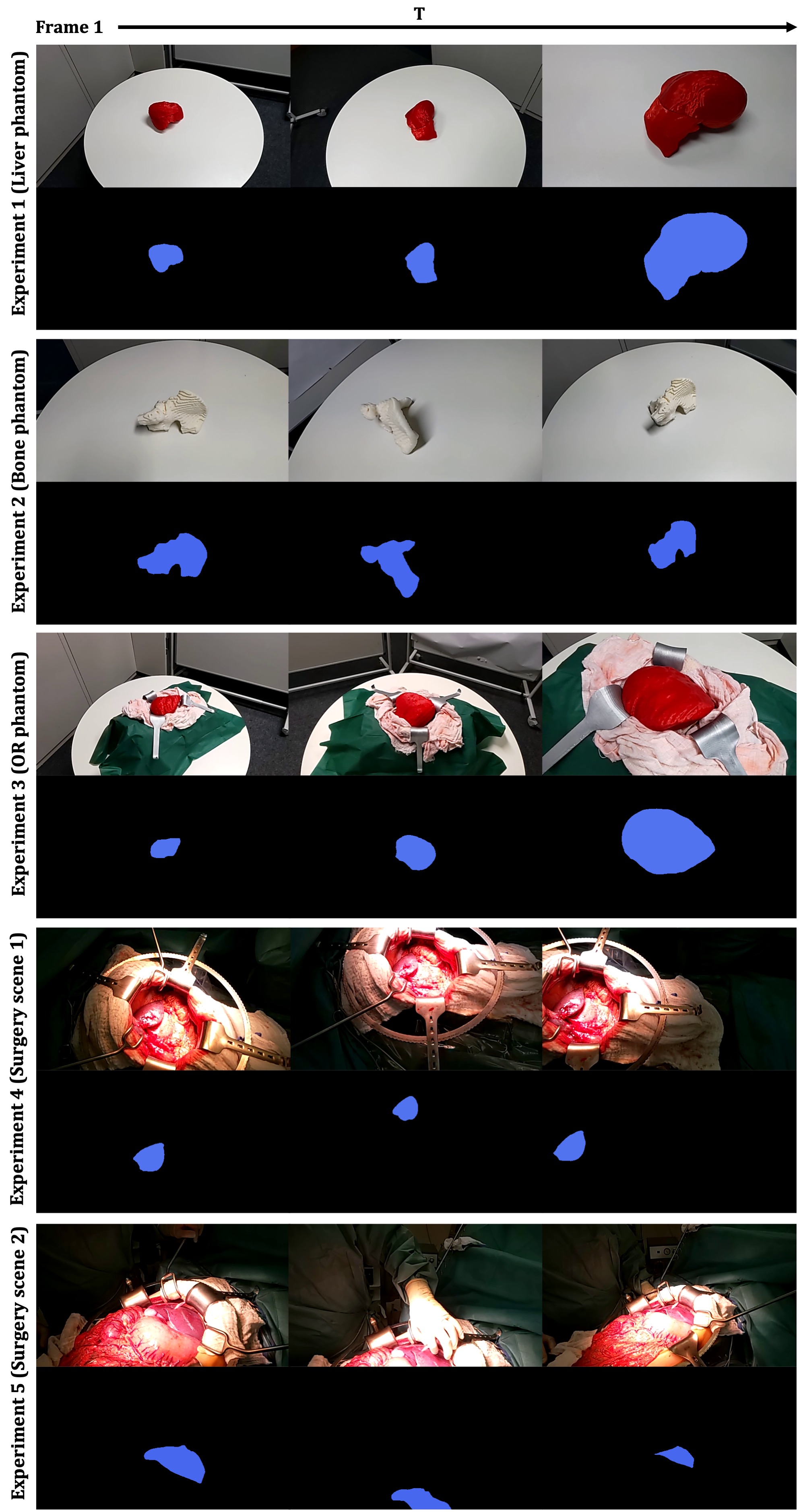}
	\caption{The figure shows three example frames (left to right) with their corresponding HOLa annotations for all five experiments.}
	\label{img:Figure2}
\end{figure}

\begin{table}
\caption{Annotation quality and time. Mean Dice scores represent the HOLa vs. human rater (HA 1) concordance for all five experiments. We further oppose the mean annotation speed (fps) of human annotators and HOLa for each experiment.}
\begin{center}
\begin{tabular}{lll}
Experiment & Annotation quality & Annotation speed \\ \midrule
& \textbf{Mean Dice} & \textbf{Annotators / HOLa} \\ \midrule
1: Liver phantom & $0.982 \pm\,0.011$ & $0.008\,fps\,/\,5\,fps$ \\
2: Bone phantom & $0.966 \pm\,0.008$ & $0.008\,fps\,/\,5\,fps$ \\
3: OR phantom & $0.981 \pm\,0.007$ & $0.008\,fps\,/\,5\,fps$ \\
4: Surgery scene 1 & $0.925 \pm\,0.013$ & $0.010\,fps\,/\,5\,fps$ \\
5: Surgery scene 2 & $0.875 \pm\,0.019$ & $0.009\,fps\,/\,5\,fps$ \\
\end{tabular}
\end{center}
\label{tab:Table1}
\end{table}

\begin{table}
\caption{Comparison of HOLa vs. human rater concordance (HA 1 vs. HOLa) to the inter-rater concordance of 5 human annotators (HA 1 vs. HA 2-5) in terms of mean Dice scores.}
\begin{center}
\begin{tabular}{lll}
Experiment & HA 1 vs. HOLa & HA 1 vs. HA 2-5 \\ \midrule
1: Liver phantom & $0.983 \pm\,0.008$ & $0.986 \pm\,0.005$ \\
2: Bone phantom & $0.967 \pm\,0.007$ & $0.967 \pm\,0.006$ \\
3: OR phantom & $0.981 \pm\,0.007$ & $0.983 \pm\,0.005$ \\
4: Surgery scene 1 & $0.921 \pm\,0.013$ & $0.939 \pm\,0.016$ \\
5: Surgery scene 2 & $0.887 \pm\,0.018$ & $0.917 \pm\,0.028$ \\
\end{tabular}
\end{center}
\label{tab:Table2}
\end{table}

Figure \ref{img:Figure2} presents HOLa annotation results for example frames of all experiments. Table \ref{tab:Table1} provides Dice scores based on 90 frames as well as a comparison of annotation speed. Table \ref{tab:Table2} opposes HOLa vs. human rater concordance to the inter-rater concordance of 5 human raters with respect to annotation quality based on a subset of 10 frames.

\begin{figure}
	\centering
    \includegraphics[width=.95\columnwidth]{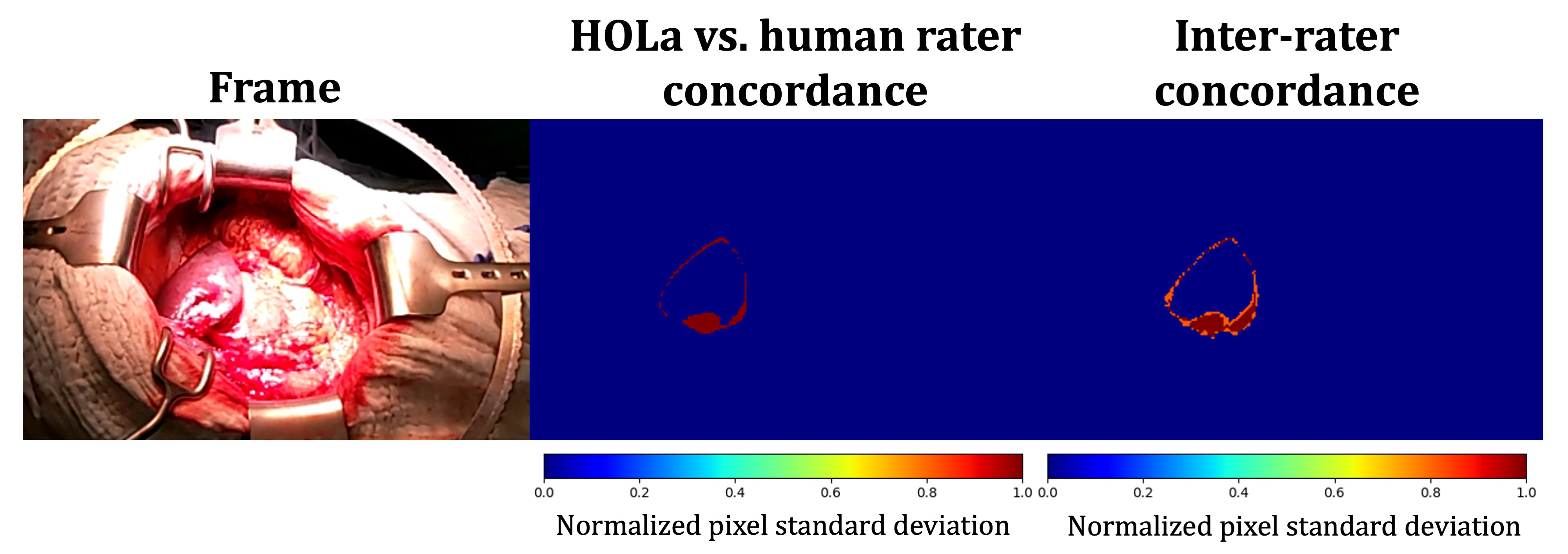}
	\caption{Annotation deviations between HOLa and human annotators for a sample clinical frame. Human annotations strongly vary at organ boundaries, especially in case of shadows and lack of color contrast between structures.}
	\label{img:Figure3}
\end{figure}

\begin{figure}
	\centering
    \includegraphics[width=.9\columnwidth]{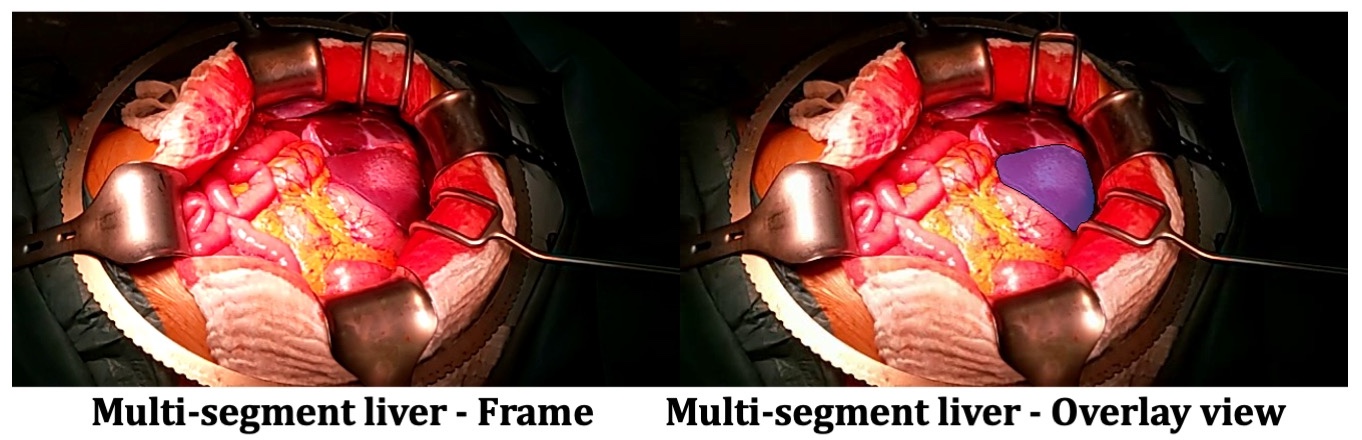}
	\caption{Applying HOLa on a liver that is divided into multiple segments leads to distortions in labeling. As a consequence, only one segment of the whole liver is labeled. However, HOLa offers to place additional seed points during quality control.}
	\label{img:Figure4}
\end{figure}

\section{Discussion}

In our preliminary results on five example labeling experiments, HOLa annotation quality was comparable to human annotations, as the HOLa vs. human rater label concordance is very close to the concordance of multiple human annotators. The label concordances are compared based on only a limited number of frames (10), however, the metrics differ by less than 0.013 Dice compared to the results based on 90 frames, suggesting that the selection is representative for the total set. As HOLa is a labeling tool, a higher Dice score means less effort for quality control on the labels. Even if technically, we can fully automate image labeling, this will never replace a human cross-check. Future work will consider this aspect more closely for performance analysis. Furthermore, since foundation models were introduced only recently, there are few studies that report on segmentation quality in comparable situations and there is none to study HoloLens data in particular.

Our work achieved Dice scores of over 0.966 on the phantom experiments. For experiment 4, which investigates a human liver clearly separated from its surroundings and with a consistent shape, the inter-rater concordance is slightly higher than the achieved mean Dice score for the corresponding HOLa vs. human rater concordance but also limited by shadows and low color contrast as demonstrated by Figure \ref{img:Figure3}. As experiment 5 investigates a liver that is strongly embedded in its surroundings with shallow transitions between the structures, we observe that the SAM model struggles to identify the exact boundaries of the liver. This consistently mitigates scores resulting in a mean Dice score of 0.875.

Since HOLa includes SAM as backbone model, SAM limitations and challenges also apply for HOLa. Specifically, SAM struggles with small and irregular objects as well as low-contrast applications like elements with similar surroundings and elements seamlessly embedded in their surroundings \citep{6}. Further problems are limited applicability and performance on professional data like medical data that SAM is not specifically trained on \citep{6}. Besides that, we experienced that it is crucial to trigger recording from a representative view on the object of interest, as an inappropriate initialization of the SAM-Track \citep{3} tracking part distorts annotation quality in downstream frames. However, in none of our experiments the target object was lost during tracking. A sequel study will be investigating the segmentation quality w.r.t a series of parameters to derive recommendations for a robust initialization.

We developed HOLa for application on single objects. This leads to incomplete segmentation if the target structure is made up of different segments. In Figure \ref{img:Figure4} for example, the surface of the liver is highly structured due to the pathology. This could be easily corrected by placing additional HOLa seed points in missing segments during the quality control. Additionally, it is important to note that distortions might occur when the object of interest disappears from the field of view. In practice, these effects are not critical for annotating individual objects, since only the frames showing the object are relevant. Furthermore, as human vision varies between HOLa users, inaccuracies might occur while setting seed points. We experienced variations of a few centimeters observing different users setting seed points at the same space. Hence, we strongly encourage users to initiate HOLa when being close to the object of interest such that there is a space of at least 2-3 centimeters between HOLa sphere cursor and object boundaries. In future work, one could further measure the ergonomics and feasibility aspect of utilizing HOLa compared to a fully computer based video labeling tool.

To our knowledge, this is the first work that evaluates the utility of the foundation model SAM on HoloLens and AR-tracking specific data. We believe this work is of value to the community since it dramatically decreases the workload of data management during the development of AI-supported HoloLens applications. Foundation models do not need to be tailored to specific data. Our solution could thus be investigated for other fields of AR research without modification.\\
\\
\textsf{\textbf{Author Statement}}\\
Research funding: Funding provided through project INMOTION by the Bavaria's State Ministry of Science and the Arts. Conflict of interest: Authors state no conflict of interest. Informed consent: Informed consent has been obtained from all individuals included in this study. Ethical approval: The research related to human use complies with all the relevant national regulations, institutional policies and was performed in accordance with the tenets of the Helsinki Declaration, and has been approved by the Ethics Committee of the Technical University Munich (DRKS00032826).

\end{document}